\documentclass[12pt,letterpaper]{article}
\usepackage[utf8]{inputenc}

\usepackage{times}
\usepackage{latexsym}
\usepackage{enumitem}
\usepackage{graphicx}
\usepackage{subfigure}
\usepackage{stmaryrd}
\usepackage{amsmath}
\usepackage[hyperref]{naaclhlt2018}
\usepackage[symbol]{footmisc}

\usepackage{titlesec}
	\titlespacing{\section}{3pt}{4pt}{4pt}
    \titlespacing{\subsection}{3pt}{4pt}{4pt}
    \titlespacing{\subsubsection}{3pt}{3pt}{3pt}

\usepackage{gb4e}\noautomath

\aclfinalcopy

\usepackage{natbib}
\title{Neural Networks and Quantifier Conservativity:\\
Does Data Distribution Affect Learnability?}
\author{
Vishwali Mhasawade$^{1}$\\\texttt{\small vishwalim@nyu.edu}
\And
Ildik\'o Emese Szab\'o$^{2}$\\\texttt{\small ildi.szabo@nyu.edu}
\And 
Melanie Tosik$^{3}$\\\texttt{\small tosik@nyu.edu}
\And
Sheng-Fu Wang$^{2}$\\\texttt{\small shengfu.wang@nyu.edu}
\AND
$^{1}$\normalfont Tandon School of Engineering\\New York University\\Two MetroTech Center\\New York, NY 11201\And
$^{2}$\normalfont Dept. of Linguistics\\New York University\\10 Washington Place\\New York, NY 10003\And
$^{3}$\normalfont Dept. of Computer Science\\New York University\\60 Fifth Avenue\\New York, NY 10011
}

\date{}

\begin{document}
\maketitle

%%% Abstract
\begin{abstract}

All known natural language determiners are conservative. Psycholinguistic experiments indicate that children exhibit a corresponding learnability bias when faced with the task of learning new determiners. However, recent work indicates that this bias towards conservativity is not observed during the training stage of artificial neural networks. In this work, we investigate whether the learnability bias exhibited by children is in part due to the distribution of quantifiers in natural language. We share results of five experiments, contrasted by the distribution of conservative vs. non-conservative determiners in the training data.
We demonstrate that the aquisitional issues with non-conservative quantifiers can not be explained by the distribution of natural language data, which favors conservative quantifiers. This finding indicates that the bias in language acquisition data might be innate or representational.

\end{abstract}

%%% Introduction
\section{Introduction}\label{intro}
\noindent Determiners in natural languages can be represented by generalized quantifiers \citep{barwiseandcooper1981gc}, which take two ordered arguments and establish a relationship between their meanings. All determiners are subject to certain cross-linguistic universals. In this paper, we focus on the conservativity universal, which states that all simple natural language determiners correspond to \textit{conservative} quantifiers \citep{barwiseandcooper1981gc,keenanstavi1986}. 

The strongest evidence for the conservativity universal emanates from first language acquisition: psycholinguistic experiments show that children are able to learn new determiners that correspond to conservative quantifiers, but not the ones corresponding to non-conservative quantifiers \citep{hunterlidz2012}. However, this finding is yet to be reproduced with neural networks --- \citet{steinertthrelkeldszymanik2017} find that neural networks can learn both conservative and non-conservative quantifiers equally well.

In this work, we pursue this issue further by conducting a series of experiments designed to investigate the role of data distribution in quantifier learning. Specifically, we examine the importance of varying the proportion of conservative and non-conservative quantifiers in the training data of a recurrent neural network (RNN).

If distributional changes in the data affect the learnability of less frequent quantifiers, it could indicate that the conservativity bias exhibited by children is a secondary effect arising from a skewed distribution. However, if changes in distribution do not affect RNN performance on learning conservative or non-conservative quantifiers, the learnability bias could be a sign of children tapping into an innate ban on non-conservative quantifiers in Universal Grammar, or of representational differences.

%%%%%%%%%%%%%%%%%%%%%%%%%%%%%%%%%%%%%%%%%%%%%%%%%%%%%%

%%% Related work
\section{Related work}\label{related-work}

%%%%%%%%%%%%%%%%%%%%%%%%%%%
%%%%% What is conservativity?
\noindent A quantifier is conservative if its second argument can be restricted to its intersection with the first argument. That is, \textit{Q(A)(B)} is conservative if its meaning does not depend on the meaning of B\textbackslash A. This definition of conservativity can be formalized as follows:

\begin{flushleft}\normalsize
\hspace{3ex}$Q$ is conservative iff:\\

\hspace{8ex}$\llbracket Q\rrbracket(A)(B)$ = $\llbracket Q\rrbracket(A)(A\cap B)$ 
\end{flushleft}

For example, $\llbracket$\textit{every}$\rrbracket$ is a conservative quantifier, because the truth conditions of the sentence in (\ref{everydoga}) are the same as that of the sentence in (\ref{everydogb}). The second argument (\textit{is brown}) can be freely replaced with the intersection of the two arguments (\textit{is a brown dog}) without changing the sentence's truth value: (\ref{everydoga}) is true only whenever (\ref{everydogb}) is true.

\begin{exe}[nosep]
	\ex Every dog is brown.\label{everydoga}
    \ex Every dog is a brown dog.\label{everydogb}
\end{exe}

%%%%%%%%%%%%%%%%%%%%%%%%%%%
%%%%% Manifestations of conservativity
\noindent The observation that all natural language determiners are conservative is supported by work on first language acquisition. \citet{hunterlidz2012} find that children can learn nonce conservative determiners, but have difficulties with non-conservative ones. In their study, a total of 20 subjects (4-5 years old) were assigned to learn a new quantifier from five training items: either the conservative \textit{gleeb}, meaning \textit{not all}, or the non-conservative \textit{gleeb'}, meaning \textit{not only}. They found that children had significant success in learning the former, but no subject in the non-conservative \textit{gleeb'} condition demonstrated having learned the new quantifier. \citet{hunterlidz2012} interpret their findings as evidence for children having a learning bias against non-conservative quantifiers.

\citet{steinertthrelkeldszymanik2017} investigate the learnability of semantic universals from a computational point of view. In their study, they train a LSTM network \cite{hochreiter1997} to perform a related sequence classification task: the input to the network is a sequence in which each element represents a set-theoretic model and a quantifier, and the output are two possible truth-values given the input. The model is a sequence of entities represented by one-hot encodings of four zones in a Venn diagram: A$\cap$B, A$\setminus$B, B$\setminus$A, and $\overline{\text{A}\cup\text{B}}$. For example, an entity in A$\setminus$B is encoded in a vector $[$0,1,0,0$]$. Two encoded quantifiers are used in each experiment: one exhibiting the semantic universal, e.g. the conservative \textit{not all} ($\vert$A$\setminus$B$\vert\neq$0), and one not exhibiting the universal, e.g. the non-conservative \textit{not only} ($\vert$B$\setminus$A$\vert\neq$0). 

While their results indicate a computational basis for other universals (monotonicity and quantity), they do not observe a similar preference for conservativity: the non-conservative quantifier \textit{not only} is not more difficult to learn than the conservative \textit{not all}. The authors acknowledge the fact that under their representation, the only difference between conservative and non-conservative quantifiers is the region of the set-theoretic model that the truth values depend on, i.e. A$\setminus$B or B$\setminus$A. Since the model is represented as a 4-dimensional vector, referencing one specific region over another is unlikely to have an effect on the learnability of the quantifier.

%SF: I feel like the following two paragraphs are out of place here. It may make sense to reformulate them and put them somewhere else. But I think if we're already running out of space, we shouldn't spend too much space on explaining LSTMs.
%Again following \citet{steinertthrelkeldszymanik2017}, we use a long short-term memory (LSTM) network \cite{hochreiter1997} for our experiments. In general, an LSTM network refers to an RNN composed of LSTM units, which in turn are composed of \textit{cells} and a number of \textit{gates} for handling the input/output. The key benefit of choosing a RNN over related architectures is the ability pass on an internal state (memory) when sequentially processing the input data.
%LSTM networks specifically have proven to be highly successful at a variety of tasks, including machine translation \citep{sutskeverVL14}, image captioning and generation \citep{vinyalsTBE14,gregorDGW15}, and question answering \citep{wang2015long}. The gating mechanisms used in LSTM units have also been shown to resemble the operation of working memory in the human brain \citep{gisigerboukadoum2011}, which makes the experimental setup biologically plausible as well.

%%%%%%%%%%%%%%%%%%%%%%%%%%%
%%%%% Sources of the bias
Therefore, we believe that the learning bias exhibited by children can have two sources: it is either innate or induced by data not encoded in \citeauthor{steinertthrelkeldszymanik2017}'s model. Linguistic explanations of the bias frequently reference innate cognitive structures. From a semantic standpoint, \citet{keenanstavi1986} argue that combining \textit{every} and \textit{some} with all extensional adjectives yields exactly the set of conservative determiners \citep[also][]{vanbenthem1986}. Other work relates this issue to the syntax-semantics interface \citep{chierchia2009,fox2002,sportiche2005,romoli2015}. \citet{fox2002} argues that trace conversion in the copy theory of movement \citep{chomsky1995} automatically results in the second argument's characteristic function being only partially defined. It is only defined for the domain of the first argument, thereby making the quantifier necessarily conservative.

Proposed explanations of the conservativity universal also posit a learnability bias. Limiting possible determiners to only conservative ones shrinks the space of possible quantifiers and their corresponding meanings, thus making the learning process faster and determiners easier to learn \citep{thijsse1985}.

The observed differences between the performance of neural networks and human learners could also stem from the differences in the experimental input representations. Neither the distribution nor the representation used by \mbox{\citeauthor{steinertthrelkeldszymanik2017}} match natural language data. While \citeauthor{steinertthrelkeldszymanik2017} train their networks on one conservative and one non-conservative nonce quantifier, children are regularly exposed to many other conservative quantifiers (recall that all natural language determiners are conservative).

\citet{steinertthrelkeldszymanik2017} encode each subset of the set-theoretic model as one input dimension. Conservative determiners are insensitive to at least one such dimension (B$\setminus$A). Therefore, an inductive bias might arise during network training, minimizing the contribution of B$\setminus$A entities if the model is trained overwhelmingly on conservative quantifiers.

Prior work investigates if and how neural networks generalize to items infrequently observed in the training data. \citet{lake2018} train RNNs to perform actions based on simplified linguistic commands (e.g. \textit{walk}, \textit{jump twice}) based on a modified version of the SCAN dataset \citep{mikolovJoulinBaroni2016}. Their results show that the RNNs' ability to generalize increases significantly when provided with a small amount of ``testing'' data. Afterwards, the network is capable of leveraging other existing combinations in the training data to produce systematic generalizations.

Moreover, \citet{linzen2016} find that just a few items can provide a good enough basis for generalizations to improve overall performance. In a set of experiments, they test the importance of ``difficulty" in subject-verb agreement sentences. They find that the model generally performs better when trained exclusively on sentences in which the nearest NP to the verb is not the one it agrees with (e.g. \textit{The \textbf{\underline{books}} on the \textbf{{shelf}} \textbf{\underline{are}} brown}), than when it is trained only on simple sentences (e.g. \textit{The \textbf{\underline{books}} \textbf{\underline{are}} brown}).

\section{Methodology}\label{methods}

\begin{table*}[ht]
\centering
 \footnotesize
\begin{tabular}{|llr|}
    \hline
    \bf{Name} & \bf{Description} & \bf{Cardinality relation}\\
    \hline\hline
    \multicolumn{3}{|l|}{\bf{Conservative quantifiers}}\\
    \hline
    \textit{all AB} & All \textit{A}s are also \textit{B}s	 & $|$A$\setminus$B$|=0$\\
   	\textit{not all AB} & Not all \textit{A}s are \textit{B}s& $|$A$\setminus$B$|\neq 0$\\
   	\textit{most AB} & Most \textit{A}s are also \textit{B}s & $\vert$A$\setminus$B$\vert ~<~\vert$A$\cap$B$\vert$\\
    \textit{most A nonB} & Most \textit{A}s are not \textit{B}s	 & $\vert$A$\setminus$B$\vert ~>~\vert$A$\cap$B$\vert$\\
    \textit{exactly half AB} & Exactly half of \textit{A}s are also \textit{B}s &$\vert$A$\setminus$B$\vert ~=~\vert$A$\cap$B$\vert$\\
    \hline\hline
    \multicolumn{3}{|l|}{\bf{Non-conservative quantifiers}}\\
    \hline
    \textit{only AB} & Only \textit{A}s are \textit{B}s (all \textit{B}s are \textit{A}s too) &$\vert$B\textbackslash$A\vert = 0$\\
   	\textit{not only AB} & Not only \textit{A}s are \textit{B}s (not all \textit{B}s are \textit{A}s too)&$\vert$B\textbackslash$A\vert\neq 0$\\
   	\textit{most BA}& Most \textit{B}s are also \textit{A}s & $\vert$B$\setminus$A$\vert ~<~\vert$A$\cap$B$\vert$\\
    \textit{most B nonA} & Most \textit{B}s are not \textit{A}s & $\vert$B$\setminus$A$\vert ~>~\vert$A$\cap$B$\vert$\\
    \textit{exactly half BA} &Exactly half of \textit{B}s are also \textit{A}s & $\vert$B$\setminus$A$\vert ~=~\vert$A$\cap$B$\vert$\\
    \hline
\end{tabular}
\caption{Complete list of quantifiers, along with descriptions and cardinality relations.} \label{table:quants}
\end{table*}

\noindent In this paper, we investigate whether or not manipulating the distribution of conservative quantifiers in the training data of a RNN can replicate the conservativity bias exhibited by human learners. If such a learning bias can be induced in neural networks, then the bias observed in children can partially be explained by the conservativity universal in natural language. If we fail to induce a learning bias, then the bias might be innate or stem from other sources, such as input representation.

We conduct a series of five experiments in which we train a LSTM network to perform a sequence classification task on artificially generated, quantificational data. The input to the network consists of a quantificational expression and a set-theoretic model representation. The quantificational expression is a one-hot encoded vector, denoting the quantifier associated with the set-theoretic model representation. The task of the LSTM network is to assign a truth value to the input pair.

The meaning of each quantifier is denoted by the relationship between two abstract sets A and B, and can refer to the various subsets of the two sets, such as the intersection or the complement of A and B, i.e. A$\setminus$B. The task of the neural network is therefore to learn which quantifier corresponds to which relationship between the two abstract sets.

%%%%%%%%%%%%%%%%%%%%%%%%%%%
%%%%% Quantifiers
\subsection{Quantifiers}\label{quants}

\noindent We define five pairs of quantifiers for a total of 10 quantifiers. Each quantifier is defined such that \textit{Q(A)(B)} = \textit{Q'(B)(A)}, where one member of each pair is conservative, and the other is not. Introducing pairs of quantifiers allows us to control for quantifier complexity. Symmetry and extensionality are also controlled for: all of our quantifiers are asymmetric, and none of them satisfy extensionality. 

In each of our five experiments, we use six quantifiers. Four of them are provided with more training data (``training" quantifiers, trained on 6,000 items each), while the remaining two quantifiers are trained on significantly less data (``testing" quantifiers, trained on 750 items each). All six quantifiers are tested on 750 items each.

The ``testing" quantifiers remain the same across all five experiments. Namely, we use the conservative \textit{all} and the non-conservative \textit{only} as ``testing" quantifiers. ``Training'' quantifiers are different in each experiment and vary in the distribution of conservative and non-conservative (C:NC) quantifiers. Specifically, we use the following distributions across the five experiments (a--e): a) 4C:0NC, b) 3C:1NC, c) 2C:2NC, d) 1C:3NC, and e) 0C:4NC. Each quantifier is represented by a 10-dimensional, one-hot encoded vector. The complete list of all 10 quantifiers and their set-theoretic denotations is given in Table \ref{table:quants}.

%%%%%%%%%%%%%%%%%%%%%%%%%%%
%%%%% Set input
\subsection{Set-theoretic input}\label{set input}
\noindent The second component of each input is a one-hot encoded, set-theoretic representation of a model. Following \citet{steinertthrelkeldszymanik2017}, we represent the model as a set of 20 entities. Each entity is a 4-dimensional, one-hot vector, denoting where the entity lies within the 4 zones of the abstract model: A$\setminus$B, A$\cap$B, B$\setminus$A, or $\overline{\text{A}\cup\text{B}}$.

For example, an entity of [1,0,0,0] is in the A$\setminus$B set, whereas an entity of [0,1,0,0] is in the A$\cap$B set. Entities of [0,0,0,0] are possible, but do not correspond to representations.

%%%%%%%%%%%%%%%%%%%%%%%%%%%
%%%%% Model architecture
\subsection{Model architecture}\label{model}

\noindent Each data point is generated by concatenating the encoded quantifier with the set-theoretic representation of 20 entities. Together, the quantifier and the model representation form a sequence of 14-dimensional vectors. The output of the LSTM are probabilities of truth values, indicating whether the input quantifier was \textit{true} or \textit{false} for the set-theoretic representation of the input entities.

The model architecture follows \citet{steinertthrelkeldszymanik2017}. Most of the hyperparameters remain unchanged to ensure comparability. The network is a standard stacked LSTM network architecture, followed by a feed-forward network to predict the output probabilities. The size of the embeddings and the mini-batches is 8, for a total of 30,000 data points (25,500 for training, 4,500 for testing). We run 3 runs of 30 trials for each experiment (90 trials in total). Our code and data are publicly available on GitHub\footnote{\url{https://git.io/vpT7x}}.

%%%%%%%%%%%%%%%%%%%%%%%%%%%%%%%%%%%%%%%%%%%%%%%%%%%%%%
%%%%% Results
 \section{Results}\label{results}
 
\begin{figure}[t]
\centering
\subfigure[Experimental condition a) (4C:0NC).]{\label{results_exp_a}\includegraphics[width=\columnwidth]{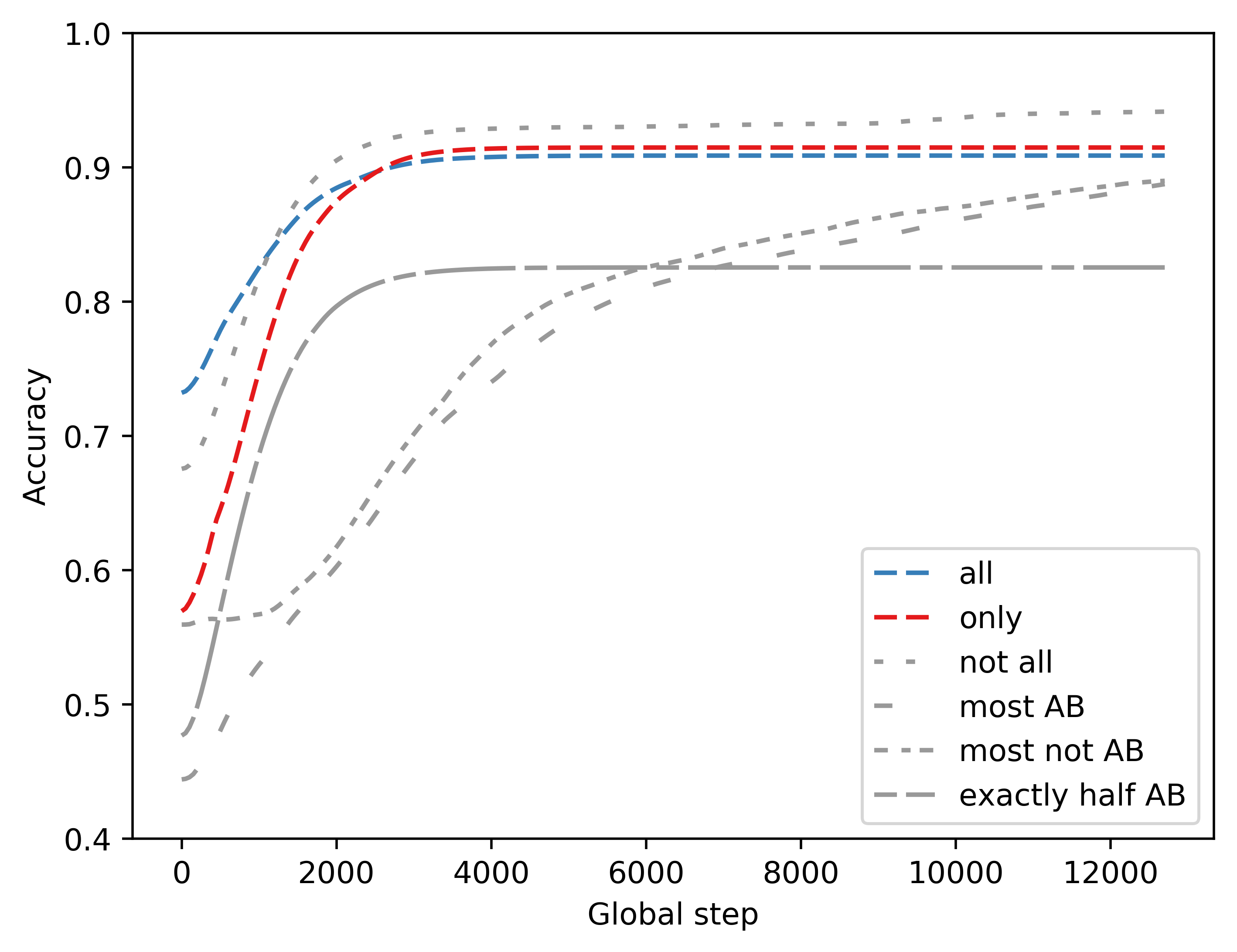}}
\subfigure[Experimental condition e) (0C:4NC).]{\label{results_exp_e}\includegraphics[width=\columnwidth]{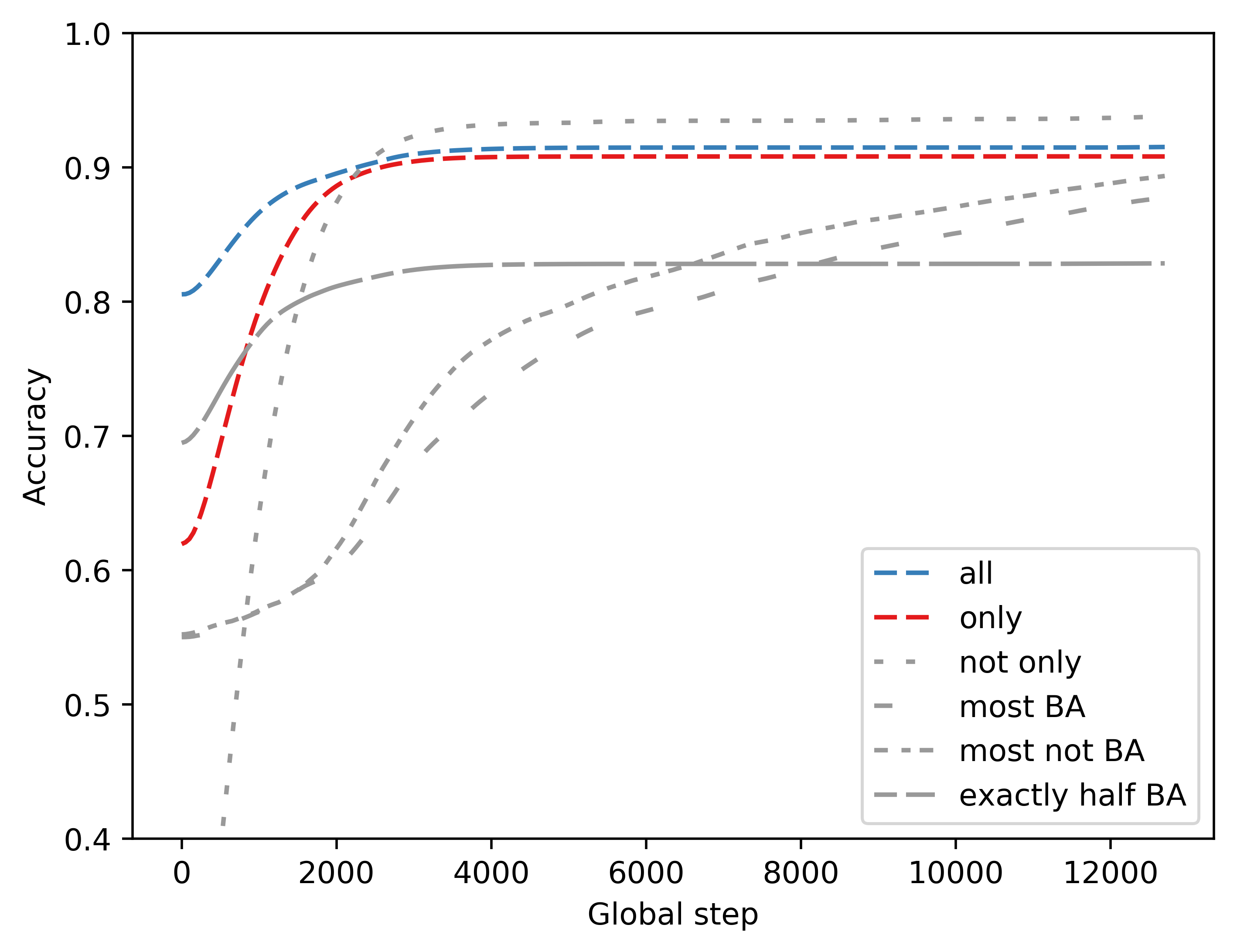}}
\caption{Median accuracy across trials by quantifier.}
\end{figure}

% \noindent Following \citeauthor{steinertthrelkeldszymanik2017}, we evaluate the learning of the conservative and non-conservative ``testing" quantifiers by comparing their convergence points during the network training phase. Convergence refers to the training step at which training and mean testing accuracy both reach and remain above 95\% until the end of each trial.
%In the end we didn't do this; we did paired t-test on accuracy instead

\noindent Both ``testing'' quantifiers (\textit{all} and \textit{only}) are learned eventually. The mean accuracy for \textit{all} at the last global step across all experiments is 91.17\%, with a median of 91.2\% and a range from 87.87\% to 95.07\%. The mean accuracy for \textit{only} at the last global step across experiments is 91.16\%, with a median of 91.2\% and a range of 88.0\% to 94.0\%.

Figure \ref{results_exp_a} shows the accuracy of the six quantifiers in experimental condition a), in which the training quantifiers are four conservative and zero non-conservative quantifiers (4C:0NC). The plot also includes the accuracies of the two testing quantifiers (\textit{all} and \textit{only}). Plotted lines reflect the median accuracies across 30 trials from one run (run 3; representative of all runs). The accuracy of the four ``training'' quantifiers is displayed in gray. Accuracies of the two ``testing'' quantifiers are shown in blue and red, where the blue line corresponds to conservative \textit{all}, and red line corresponds to non-conservative \textit{only}.

Figure \ref{results_exp_e} contains results for the opposite data distribution. Specifically, it illustrates the accuracies for experiment e), corresponding to condition 0C:4NC, in which the distribution of ``training'' quantifiers was zero conservative and four non-conservative quantifiers. Both plots visually indicate that there is no noticeable difference between the accuracy on the two ``testing" quantifiers.

In order to confirm that the results on the ``testing'' quantifiers are systematic, we conduct a paired t-test on the quantifier accuracies at every 50 steps of training. Since we are testing multiple hypotheses at once, we use the Bonferroni correction method to compensate for the increased likelihood of false positives.  

Starting from a desired overall alpha level $\alpha_0=0.05$, we need to correct for the number of hypotheses, $m$. $m$ is equal to our number of experiments multiplied by the number of steps we evaluate for each experiment. Since the convergence point for each experiment lies at approx. 3001 steps, and training accuracies are evaluated at every 50 steps, we evaluate a total of 61 steps (from global step 1 to 3001) for each of the five experiments. Therefore, the number of hypotheses $m = 61 * 5 = 305$.

Using Bonferroni correction, we test each individual step at a statistical significance level of $\alpha = 0.05/305 = 0.000164$. After correcting the threshold of statistical significance according to our experimental setup, we find that there is no significant difference between the accuracy of \textit{all} and \textit{only} at any point during the first 3001 steps of model training, across all experiments. 

While all quantifiers are learned eventually, Figure \ref{results_exp_a} and \ref{results_exp_e} suggest that there are inherent differences among the ``training" quantifiers. For example, the quantifier pair of \textit{not all} and \textit{not only} appears to be easier for the network to learn than \textit{most AB} and \textit{most BA}.

Therefore, we run two additional paired t-tests for two new quantifier pairings: one new conservative pair, \textit{most AB} and \textit{not all}, and one new non-conservative pair, \textit{most BA} and \textit{not only}. Both pairs occur in three of the five experiments. Thus, for each pair, the alpha level is adjusted by 183 hypotheses (3 $\times$ 61), $\alpha/m = 0.05/183 =0.000273 $. Results show that the accuracies for \textit{not all} and \textit{not only} are significantly higher than the accuracies of \textit{most AB} and \textit{most BA}, respectively.

%%%%%%%%%%%%%%%%%%%%%%%%%%%%%%%%%%%%%%%%%%%%%%%%%%%%%%
%%%%% Analysis
 \section{Analysis}\label{analysis}

\noindent Overall, our results show that manipulating the distribution of conservative quantifiers in the training data of an LSTM network does not induce a learning bias towards conservativity. Quantifier learning for conservative quantifiers does not converge faster when the learning data is biased towards conservative (or non-conservative) quantifiers. \textit{All} and \textit{only} are learned to an equal extent and at the same rate across all five experimental conditions. In the current experimental setup, the network does not learn to ignore one particular zone in the Venn diagram (corresponding to one dimension in the input vectors). In other words, the network does not on its own develop a dispreference for non-conservative quantifiers.

Furthermore, we argue that the observed differences among the ``training" quantifiers can be explained by the effect of quantifier complexity. For both of the quantifiers \textit{most AB} and \textit{most BA}, the truth value of the expression depends on the relation between the following two cardinalities: $\vert$A$\setminus$B$\vert$ and $\vert$A$\cap$B$\vert$. Verifying the cardinality relations for \textit{most AB} and \textit{most BA} is therefore significantly more complex than evaluating the cardinality of \textit{not all} or \textit{not only}, which only concerns $\vert$A$\setminus$B$\vert$ or $\vert$B$\setminus$A$\vert$, respectively.

Since all quantifiers are symmetrically distributed across experimental conditions (using their pairwise equivalence), we are confident that our results are robust and in no way influenced by a confound in quantifier complexity. The learnability of the ``testing'' quantifiers is not impacted by the distribution of conservative vs. non-conservative quantifiers in the training data.

%%%%%%%%%%%%%%%%%%%%%%%%%%%%%%%%%%%%%%%%%%%%%%%%%%%%%%
%%% Conclusion
\section{Conclusion and future work}\label{conc}

\noindent In this paper, we demonstrate that the conservativity of ``training'' quantifiers does not have any meaningful effect on the learnability of conservative and non-conservative ``testing" quantifiers. Varying the distribution of conservative and non-conservative quantifiers in the training data of a LSTM network does not result in a measurable difference between quantifier accuracies across five experimental conditions. Training the model on only conservative quantifiers does not facilitate the learning of conservative quantifiers at test time. While we do observe an effect related to quantifier complexity, this effect is explicitly controlled for in the experimental conditions, and therefore has no impact on the effects of training data distribution. 

Our results indicate that the children's learnability bias favoring conservative quantifiers is not merely an artifact of the distribution of quantifiers in natural language. Thus, we believe that the learnability bias must be reflective of either a more general, innate bias against conservative quantifiers, or due to additional factors that differ between the simulated and experimental learning situations.

In order to verify the latter possibility, further works needs to be done. One challenging avenue of future work would be to investigate the differences in the input representations. \citet{steinertthrelkeldszymanik2017} reduce the meaning of quantifiers to abstract relationships between sets, which does not accurately reflect the order of the quantifier arguments that correspond to the two sets. Enhancing the existing representations to incorporate quantifier arguments as well could bring the experimental setup one step closer to natural language data.

\bibliography{termpaper}

\end{document}